%% file: iclr2022_conference.tex
\documentclass{article} 
\usepackage{iclr2022_conference,times}

\input{math_commands.tex}

\usepackage{hyperref}
\usepackage{url}
\usepackage[pdftex]{graphicx}

\title{A weakly supervised framework for \\ high-resolution crop yield forecasts}

\author{Dilli R. Paudel, Diego Marcos, Allard de Wit, Hendrik Boogaard \& Ioannis N. Athanasiadis \\
Wageningen University and Research \\
PO Box 47, 6700 AA Wageningen, The Netherlands. \\
\texttt{\{dilli.paudel,diego.marcos,allard.dewit,} \\ \texttt{hendrik.boogaard,ioannis.athanasiadis\}@wur.nl} 
}

\iclrfinalcopy 
\begin{document}

\maketitle

\begin{abstract}
Predictor inputs and label data for crop yield forecasting are not always available at the same spatial resolution. We propose a deep learning framework that uses high resolution inputs and low resolution labels to produce crop yield forecasts for both spatial levels. The forecasting model is calibrated by weak supervision from low resolution crop area and yield statistics. We evaluated the framework by disaggregating regional yields in Europe from parent statistical regions to sub-regions for five countries (Germany, Spain, France, Hungary, Italy) and two crops (soft wheat and potatoes). Performance of weakly supervised models was compared with linear trend models and Gradient-Boosted Decision Trees (GBDT). Higher resolution crop yield forecasts are useful to policymakers and other stakeholders. Weakly supervised deep learning methods provide a way to produce such forecasts even in the absence of high resolution yield data.
\end{abstract}

\section{Introduction}
\label{intro}

Predictor inputs and label data for crop yield forecasting are often not available at the same spatial resolution. Label data, such as yield statistics, are published at regional and national level. Weather inputs are available at grid-level \citep{ECJRC2022, Daymet2020} and soil and remote sensing data at sub-kilometer resolutions \citep{ESDAC2021, poggio_etal2021, Copernicus2022}. Common statistical and machine learning methods require both inputs and labels at the same spatial level. Therefore, predictor inputs are usually aggregated to the level of yield data. Deep learning methods can handle input and label data at two spatial levels, limiting the spatial aggregation required for input data. Neural network architectures can be trained using high resolution inputs and low resolution yield data to produce crop yield forecasts for both spatial levels.

Many studies have used deep learning for crop yield forecasting \citep{fan_etal2021, shahhosseini_etal2021, wolanin_etal2020, khaki_etal2020}, but they do not disaggregate yields to high resolutions. Other methods of disaggregating crop yields exist, for example, area-to-point kriging \citep{brus_etal2018, steinbuch_etal2020} and spatial allocation based on cross-entropy method \citep{you_etal2014} or remote sensing indicators \citep{shirsath_etal2020}. We draw inspiration from \citet{jacobs_etal2018}, who trained a convolutional neural network and an aggregation layer to predict pixel-level population density from high resolution satellite images and low resolution density statistics. To our knowledge, weakly supervised methods have not been used to disaggregate crop yields to high resolutions.

We propose a weakly supervised deep learning framework that uses high resolution inputs  and low resolution labels to produce crop yield forecasts for both spatial resolutions. Predictor inputs come from NUTS3 regional level and yield and crop area statistics from NUTS2 level, with NUTS2 and NUTS3 regions representing the low and high spatial resolutions. NUTS (Nomenclature of Territorial Units for Statistics) is a hierarchical system of dividing the territory of the European Union for statistics and policy \citep{Eurostat2016}. Our approach is weakly supervised because models to produce NUTS3 forecasts are trained using low resolution NUTS2 labels. Our objective is to build and evaluate crop yield forecasting models that can produce high resolution forecasts even when high resolution yield data are not available. This objective can be viewed from two perspectives. First, we assess the performance of weak supervised models in disaggregating crop yields from low to high resolution. Second, we evaluate the quality of low resolution yield forecasts produced using high resolution inputs. Our analysis includes two crops (soft wheat and potatoes) and five countries: Germany (DE), Spain (ES), France (FR), Hungary (HU) and Italy (IT).

The rest of the paper is structured as follows: Section \ref{methods} describes data and methods; Section \ref{results} presents the results; and Section \ref{discussion} discusses our findings and outlines directions for future work.

\section{Methods}
\label{methods}

Our weakly supervised deep learning framework uses high resolution predictor inputs and low resolution labels to produce crop area and yield forecasts for both spatial resolutions. To evaluate these yield forecasts, we compared performance with linear trend models and Gradient-Boosted Decision Trees. To assess their usefulness to policymakers and other stakeholders, we analyzed the spatial variability of high resolution forecasts.

\subsection{Data}
Data sources used in this paper are summarized in Table A.1. Most of our data came from the MARS Crop Yield Forecasting System of the European Commission's Joint Research Centre \citep{MARSWiki2021} and covered two crops and five countries: soft wheat (DE, ES, FR, IT) and potatoes (DE, FR, HU, IT). Data from all countries was combined to build one prediction model per crop. Seasonal data included outputs of the WOFOST crop model \citep{vanDiepen_etal1989, supit_etal1994, deWit_etal2019}, weather variables and remote sensing indicators aggregated to NUTS3. The yield trend was captured using yield values of five previous years at NUTS2. Static differences among regions were captured by soil water holding capacity and agro-environmental features, such as elevation, slope, field sizes, irrigated area \citep{paudel_etal2022}. In addition, agro-environmental zones and countries were added as categorical variables to account for other agro-climatic and administrative differences. Reported yield and crop area statistics at NUTS2 served as labels. In most cases, we had data from 1999 to 2018. The most recent 30\% of the years were allocated to the test set. From the remaining 70\% training years, 5 most recent years were used for a sliding-window 5-fold validation (Figure A.1).

\subsection{The Weakly Supervised Framework}

\begin{figure}[h]
\begin{center}
\includegraphics[width=0.9\linewidth]{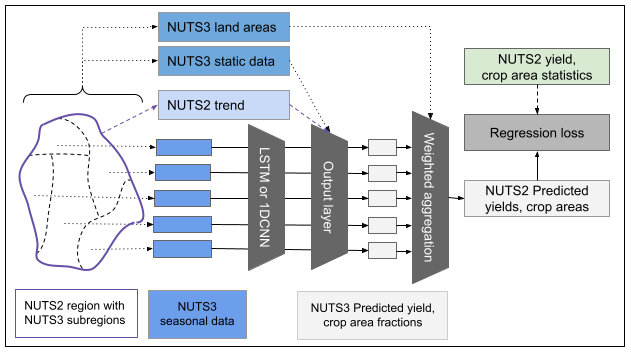}
\end{center}
\caption{Weakly supervised framework to produce high resolution crop yield forecasts.}
\end{figure}

We selected two neural network architectures, namely long short-term memory (LSTM) networks and 1-dimensional convolutional neural networks (1DCNN), that can learn from seasonal time series of predictors. Both LSTMs and 1DCNNs have been used in literature to learn features from sequential data \citep{you_etal2017, khaki_etal2020}. Seasonal data at NUTS level 3, including crop model outputs, weather and remote sensing indicators, was processed by LSTM or 1DCNN. Features from the LSTM or 1DCNN layers, together with static agro-environmental data and yield trend features (based on NUTS2 yields), were passed to an output layer (Figure 1). An aggregation layer computed crop area weights and aggregated the forecasts to low resolution (NUTS2). The framework was supervised with NUTS2 yields and crop areas. Data from all NUTS3 regions within a NUTS2 region formed a batch to enable aggregation of NUTS3 forecasts.

The weakly supervised model (WS model) produced NUTS3 yield forecasts and crop area fractions. We believe remote sensing inputs can predict crop area fractions (crop area/land area), but not the absolute crop areas. Predicted crop area fractions were multiplied with land areas to produce NUTS3 crop areas, which were used to calculate crop area weights for the aggregation layer.

\subsection{Evaluation}

\begin{figure}[h]
\begin{center}
\includegraphics[width=0.7\linewidth]{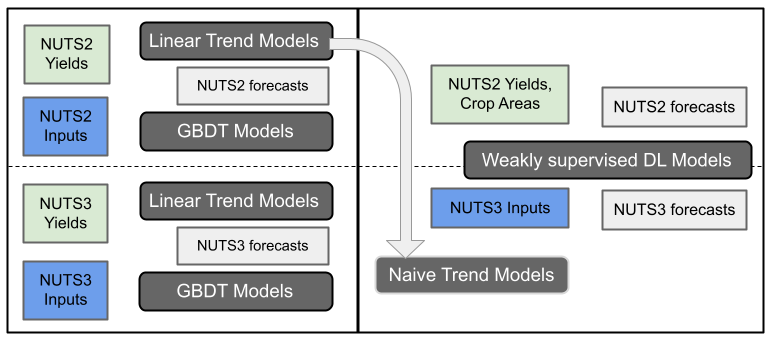}
\end{center}
\caption{Evaluation framework.}
\end{figure}

Forecasts from the WS model were evaluated at both spatial resolutions. The quality of disaggregated NUTS3 yields was assessed by comparing with two sets of models. The first set was a naive trend model that acted as our baseline for disaggregation. The naive trend model predicted the NUTS2 yield trend as NUTS3 forecasts for all NUTS3 sub-regions. The second set included the NUTS3 trend model and Gradient-Boosted Decision Trees (GBDT) model (Figure 2). Unlike the naive trend and WS models, the NUTS3 trend and GBDT models had access to NUTS3 yield data. The NUTS3 GBDT and WS models used the same predictor inputs, except yield trend features: GBDT had access to NUTS3 trend; WS only had NUTS2 trend. Forecasts were made 60 days before harvest. At NUTS2, WS model forecasts were compared with the NUTS2 trend model and NUTS2 GBDT model. All trend models were calibrated with yield values of five previous years.

Models were compared based on box plots of prediction residuals (i.e., predicted yield - reported yield) and normalized root mean squared errors (NRMSE), normalized by average yield of the test set. The significance of model performance was evaluated by running Wilcoxon signed-rank test on the prediction residuals  of different models.

In addition to error comparisons, we also analyzed the spatial variability of NUTS3 yield forecasts from disaggregation models (naive trend and WS model). A significant part of yield variability is explained using the yield trend attributed to factors such as technological improvements (see \citet{lecerf_etal2019}). We expected the NUTS2 trend features to make the model more accurate, but suppress spatial variability among NUTS3 regions. Therefore, we ran a version of the WS model without NUTS2 trend to learn spatial differences. Spatial variability of NUTS3 yield forecasts was analyzed for soft wheat (FR). We selected regions based on maximum acreage and years (2016 and 2017) based on significant yield losses reported in the north of FR in 2016 (see \citet{benari_etal2018}).

\section{Results}
\label{results}
In this section, we report results from the LSTM version of the WS model because of its superior validation set performance over the 1DCNN one (Figure A.2).

\subsection{Evaluation of high resolution yield forecasts}
The WS models were statistically similar to NUTS3 GBDT models (p-values: 0.67 for soft wheat and 0.265 for potatoes) (Table A.2), and significantly better than the naive trend models (p-values near zero). For soft wheat, the WS model was also better than the NUTS3 trend model (p-value near zero). For soft wheat, interquartile range was smaller and per-country NRMSEs were mostly lower compared to both trend models (Figure 3). Disaggregation worked less well for potatoes, but the performance was still better than the naive trend model (Table A.2, Figure A.3).

\begin{figure}[h]
\begin{center}
\includegraphics[width=0.5\linewidth]{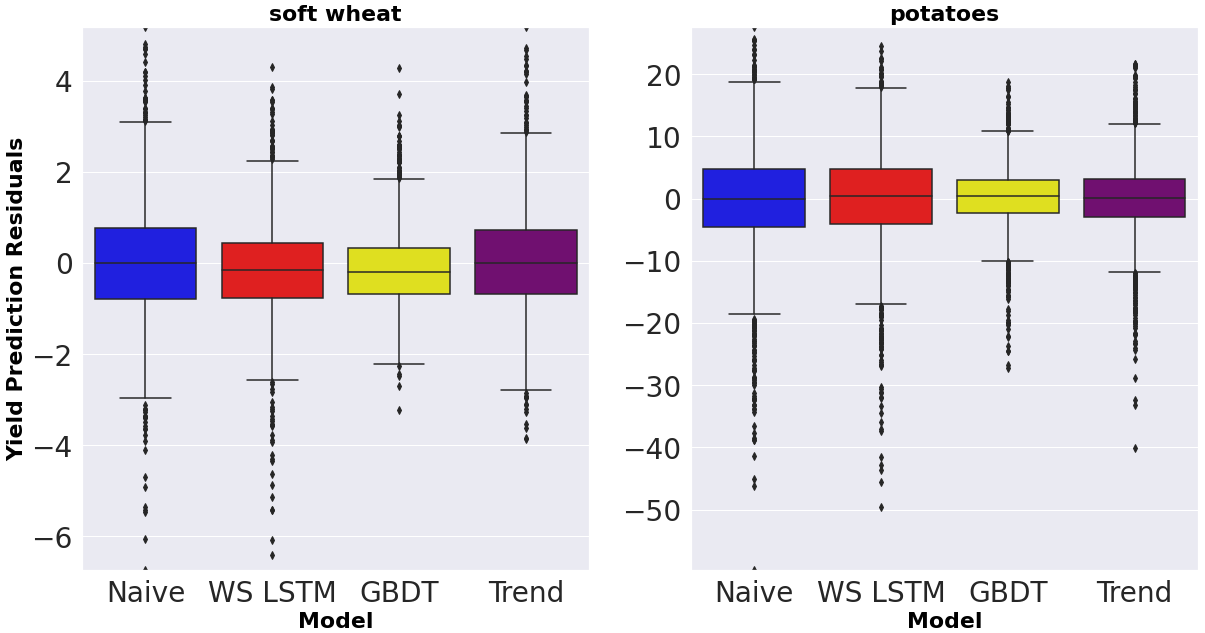}
\includegraphics[width=0.45\linewidth]{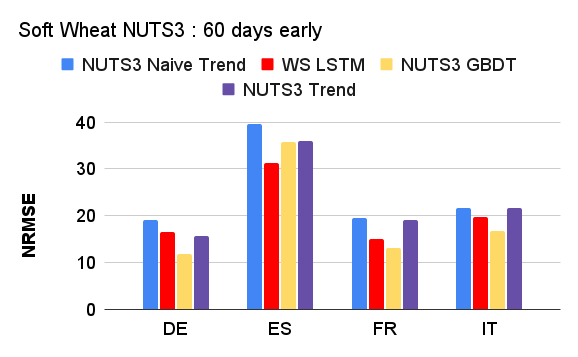}
\end{center}
\caption{\textit{Left:} Boxplots of NUTS3 prediction residuals. \textit{Right:} NRMSEs for soft wheat.}
\end{figure}

\subsection{Evaluation of low resolution yield forecasts}
The WS models produced NUTS2 forecasts that were similar compared to NUTS2 GBDT models (p-values: 0.083 for soft wheat and 0.455 for potatoes) (Table A.3), and better than the NUTS2 trend models (p-values: 0.00095 for soft wheat and 0.000002 for potatoes). Box plots and per-country NRMSEs again showed that residuals and errors were smaller for soft wheat than for potatoes (Figure 4, A.3). The WS models had smaller median residuals compared to GBDT models for both crops (Table A.3), indicating some value in using NUTS3 inputs to forecast at NUTS2.

\begin{figure}[h]
\begin{center}
\includegraphics[width=0.5\linewidth]{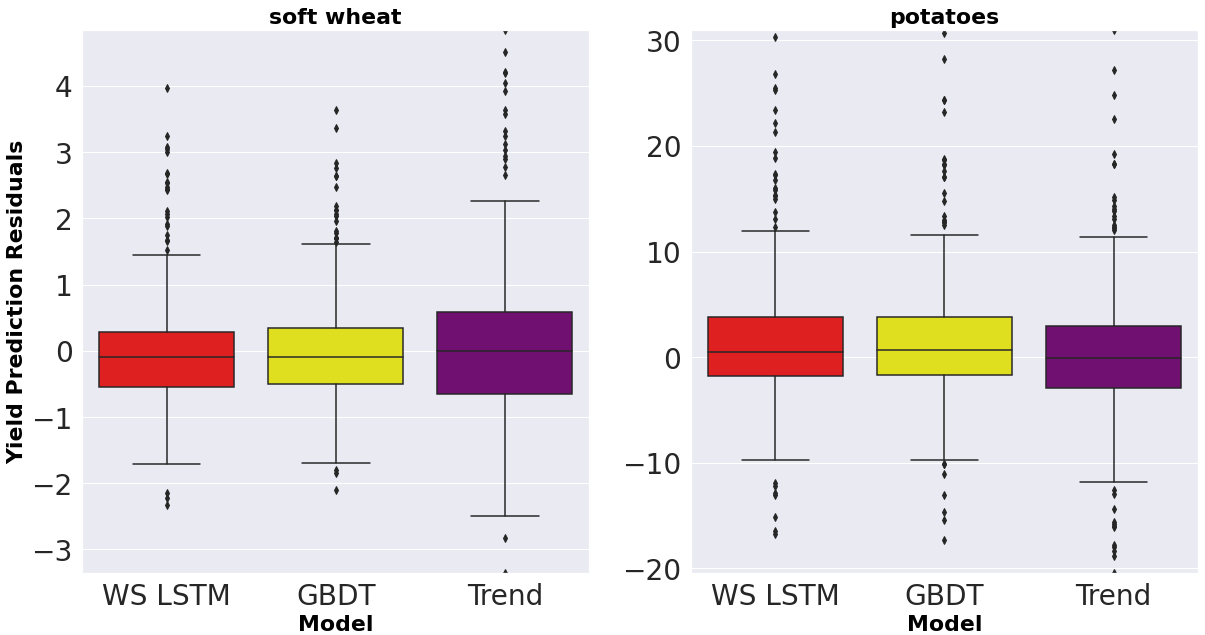}
\includegraphics[width=0.45\linewidth]{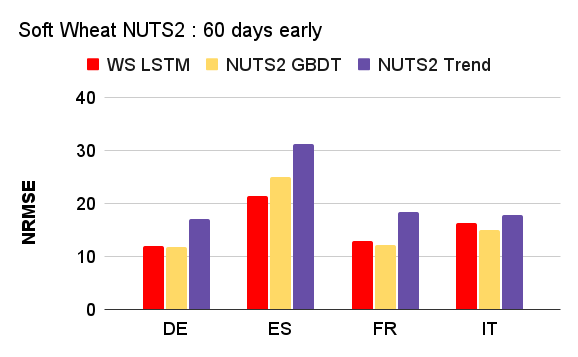}
\end{center}
\caption{\textit{Left:} Boxplots of NUTS2 prediction residuals. \textit{Right:} NRMSEs for soft wheat.}
\end{figure}

\subsection{Spatial variability of high resolution forecasts}
As mentioned before, the north of France had significant yield losses for soft wheat in 2016. The naive trend model predicted higher yields, with an average prediction residual of 2.865. The WS model predicted slightly lower (average residual: 2.21) but still higher than the reported yields (Figure 5, \textit{top}). In 2017, the naive trend model was heavily influenced by the 2016 yields (especially in the middle: FRI3, FRB0, FRC1), while the WS model predicted values similar to the reported yields (Figure 5, \textit{bottom}). As expected, the WS model with trend did not capture yield variability within NUTS2 regions. The WS No Trend model captured such differences better, but generally underestimated the yields: the average prediction residual was -0.57 compared to -0.028 for the WS model using trend. Combining information from both versions of the WS model provided more accurate estimates of yields as well as spatial differences among NUTS3 regions.

\begin{figure}[h]
\begin{center}
\includegraphics[width=0.9\linewidth]{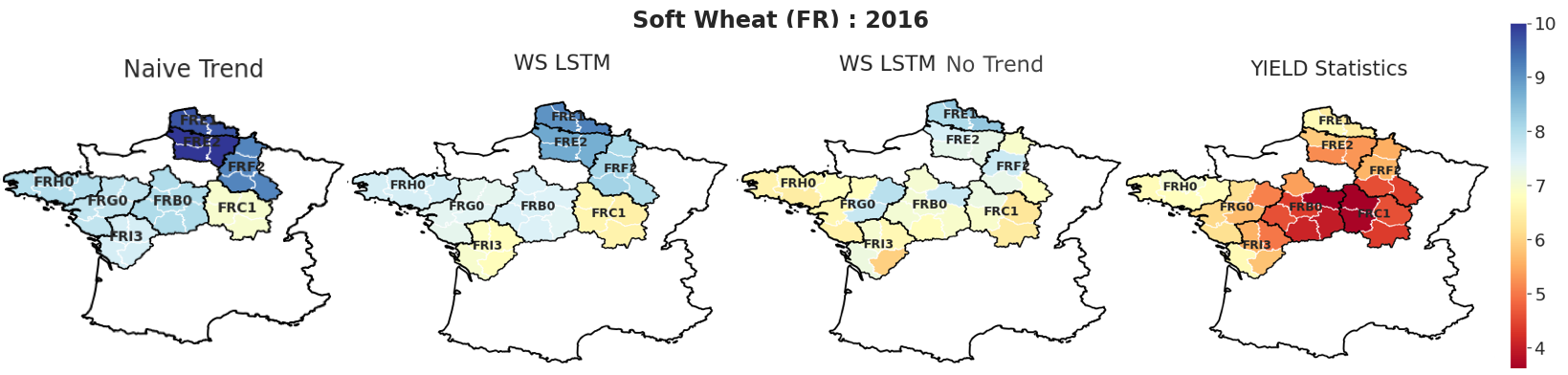}
\includegraphics[width=0.9\linewidth]{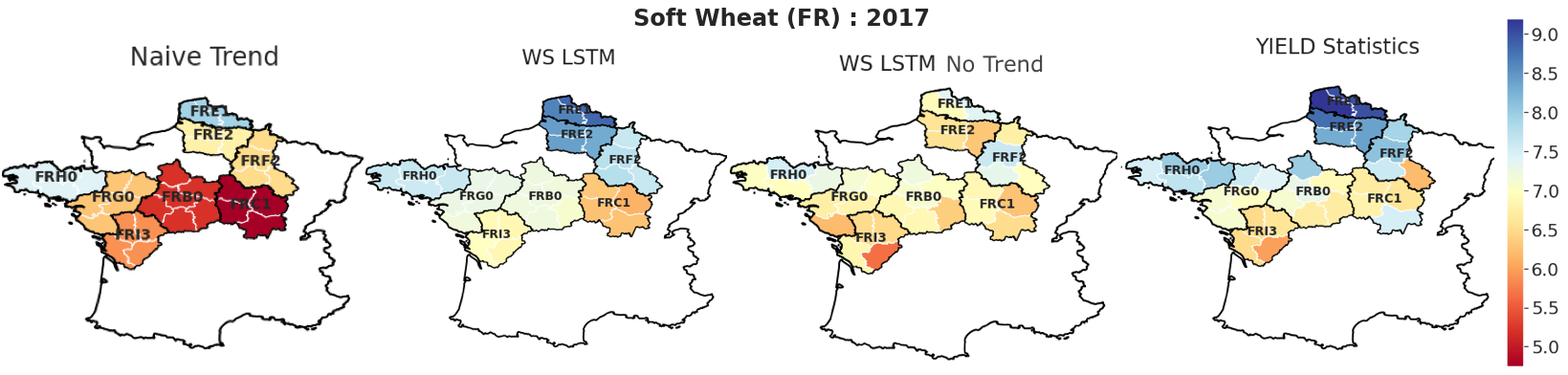}
\end{center}
\caption{Spatial variability of soft wheat yields and forecasts. \textit{Top:} FR 2016. \textit{Bottom:} FR 2017}
\end{figure}

\section{Discussion}
\label{discussion}

Data on crop yield predictors will be increasingly available at higher resolutions. Yield data may not be available due to many reasons, including privacy concerns. When there is an imbalance between spatial resolutions of inputs and yields, weak supervised methods provide a solution. Our results showed that weakly supervised models were able to produce reliable NUTS3 yield forecasts, especially for soft wheat, without using NUTS3 yields. The no-trend version of the model also captured some of NUTS3-level yield variability. Our approach will continue to work when yield data is available at low resolution for some regions and at high resolution for others. Similarly, high resolution crop areas, when available, will further improve the quality of yield forecasts.

In this paper, we have scratched the surface of high resolution crop yield forecasting without strong supervision. We see three areas that need further research to gauge the ability of weak supervision to produce reliable forecasts. First, the scale differences that can be handled by weak supervision needs investigation. For example, weak supervision from NUTS3 yields may produce good quality forecasts for 25km grids, but not 1km grids. Second, predictor inputs must be suitable to capture yield variability at selected resolutions. Weather variables may influence NUTS2 and NUTS3 yields, but become less relevant at farm level. Third, we experimented with standard neural network architectures. Future work could investigate other architectures that are more suitable for weak supervision. Data size and quality will always play a role due to the data-driven nature of neural networks.

High resolution crop yield forecasts provide useful information to policymakers and other stakeholders for local analysis and monitoring. We have shown that weakly supervised methods can produce such forecasts in the absence of high resolution labels.

\section{Acknowledgments}
This work was partially supported by the European Union’s Horizon 2020 research and innovation programme under grant agreement No. 825355 (CYBELE). We would like to thank S. Niemeyer from the European Commission’s Joint Research Centre (JRC) for permission to use MARS Crop Yield Forecasting System data. We also would like to thank M. van der Velde, L. Nisini and I. Cerrani from JRC for sharing the Eurostat regional yield statistics and crop areas.

\bibliography{iclr2022_conference}
\bibliographystyle{iclr2022_conference}

\pagebreak
\appendix
\setcounter{figure}{0}
\setcounter{table}{0}
\renewcommand{\thefigure}{A.\arabic{figure}}
\renewcommand{\thetable}{A.\arabic{table}}
\section{Appendix}
\subsection{Data Sources}
Most of our data comes from the MARS Crop Yield Forecasting System (MCYFS) of European Commission’s Joint Research Centre \citep{MARSWiki2021}. Other sources are indicated in Table A.1.

\begin{table}[!ht]
\caption{Data sources summary}
\label{data-sources}
\begin{center}
\begin{tabular} { | p{0.2\linewidth} | p{0.7\linewidth} |}
\multicolumn{1}{c}{\bf DATA}  &\multicolumn{1}{c}{\bf INDICATORS, SOURCES}
\\ \hline
WOFOST crop model outputs & Water-limited dry weight biomass ($kg$ $ha^{-1}$), Water-limited dry weight storage organs ($kg$ $ha^{-1}$), Water-limited leaf area divided by surface area ($m^2$ $m^{-2}$), Development stage ($0-200$), root-zone soil moisture as \% of soil water holding capacity, sum of water limited transpiration ($cm$). \textbf{Source}: MCYFS. See \citet{lecerf_etal2019}. \\ \hline
Meteo & Maximum, minimum, average daily air temperature (\textdegree C), sum of daily precipitation (PREC) ($mm$), sum of daily evapotranspiration of short vegetation (ET0) (Penman-Monteith, \citet{allen_etal1998}) ($mm$), climate water balance = (PREC - ET0). \textbf{Source}: MCYFS. See \citet{lecerf_etal2019}. \\ \hline
Remote Sensing & Fraction of Absorbed Photosynthetically Active Radiation (Smoothed) (FAPAR).
\textbf{Source}: MCYFS. See \citet{Copernicus2020}. \\ \hline
Crop Areas & Absolute crop areas ($ha$). Fraction of parent regions’s crop area. \textbf{Source}: Eurostat \citep{Eurostat2021a} and MCYFS \citep{ECJRC2022}. \\ \hline
Irrigated area & Irrigated total area and irrigated crop-specific area. \textbf{Source}: \citet{ECJRC2022}. \\ \hline
Elevation, slope & Average and standard deviation. \textbf{Source}: \citet{USGSEROS2021}. \\ \hline
Soil & SM\_WHC (water holding capacity). \textbf{Source}: MCYFS. See \citet{lecerf_etal2019}. \\ \hline
Field Size & Average and standard deviation. \textbf{Source}: \citet{lesiv_etal2019}. \\ \hline
Yield & Yield at NUTS3 level. \textbf{Source}: \citet{FRAgreste2020, DERegionalStats2020, Eurostat2021a, ECJRC2022}. \\ \hline
\end{tabular}
\end{center}
\end{table}

\subsection{Weakly Supervised Model}
The weakly supervised model was supervised using NUTS2 crop areas and yields. The combined loss was the sum of the two losses (crop area loss and yield loss) normalized by the training set average of the corresponding labels. The hyperparameters learning rate and L2-penalty lambda were optimized using custom sliding validation (Figure A.1; \citet{paudel_etal2022}). After optimizing hyperparameters, the model was trained on the entire validation set (no 5-fold) with early stopping: training stopped after validation error increased for two successive epochs. The optimized hyperparameters and early stopping epochs were used to evaluate the model on the test set.

\subsubsection{1DCNN Architecture}
CNN Layer1 : Conv1d(11, 16, kernel\_size=(3,), stride=(1,), padding=(1,))
\\
CNN Layer 2: Conv1d(16, 32, kernel\_size=(3,), stride=(2,), padding=(1,))
\\
CNN Layer 3: Conv1d(32, 8, kernel\_size=(3,), stride=(2,), padding=(1,))
\\
Batch Normalization, ReLU Activation and Dropout(p=0.1) were added after each CNN layer.
\\
Output Layer: Linear(in\_features=100, out\_features=2, bias=True)

\subsubsection{LSTM Architecture}
LSTM Layer : LSTM(11, 64, batch\_first=True)
\\
Output Layer : Linear(in\_features=100, out\_features=2, bias=True)

Both LSTM and 1DCNN were implemented using pytorch (https://pytorch.org/).

\subsection{GBDT Model}
Input data and training, validation and test splits were identical between the GBDT model and the weakly supervised model, except for trend features (NUTS3 vs NUTS2). The GBDT model used in this paper has some differences compared to \citet{paudel_etal2022}. First, a combined model was built for four countries. Second, agro-environmental zones and countries were added as categorical features.

The GBDT model is based on GradientBoostingRegressor() from scikit-learn \citep{pedregosa_etal2011}. Hyperparameters including GBDT parameters, feature selector and number of features optimized using BayesSearchCV from scikit-optimize package \citep{ScikitOptimize2021}. Feature selectors included a Random Forests model (SelectFromModel) and Recursive Feature Elimination using a Lasso Regression model.

\subsection{Training, validation and test splits}
\begin{figure}[h]
\label{validation-splits}
\begin{center}
\includegraphics[width=0.75\linewidth]{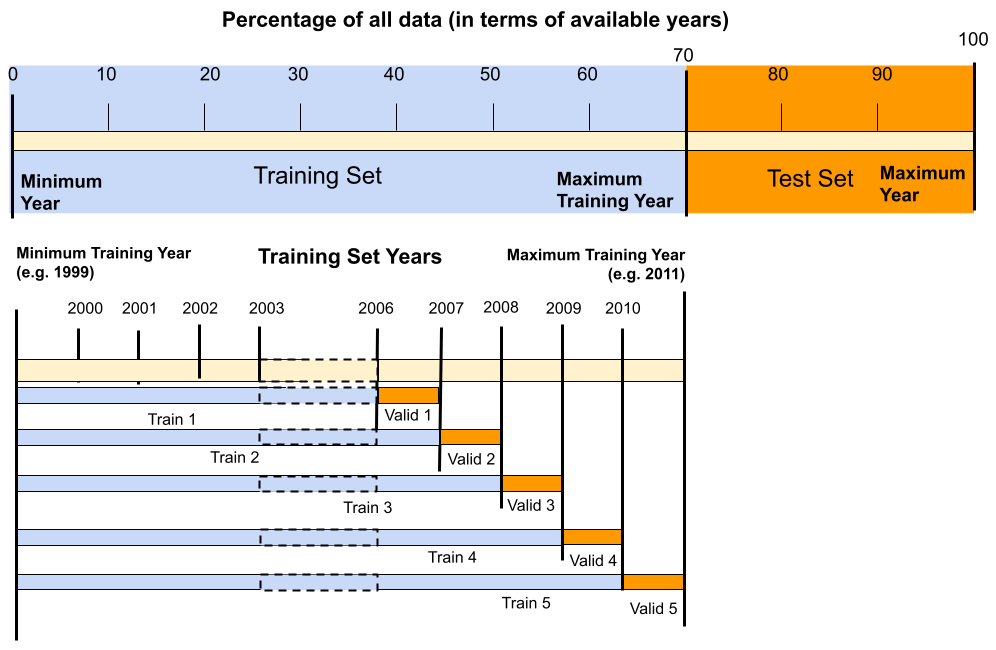}
\end{center}
\caption{Training, validation and test splits. Reproduced from \citet{paudel_etal2022}}
\end{figure}

\subsection{Software implementation}
Software implementation of the weakly supervised framework can be accessed here:
\\
https://github.com/BigDataWUR/MLforCropYieldForecasting/tree/weaksup

\subsection{Validation set results for Architecture Selection}
The decision to use combined data from four countries was based on validation set performance of strongly supervised NUTS3 models for soft wheat. Since NRMSEs were similar for both cases, we chose to use combined data because the larger data size would limit overfitting issues. CV comparisons for weak supervision using 1DCNN and LSTM showed that LSTM had lower NRMSEs on the validation set (Figure A.2).

\begin{figure}[h]
\begin{center}
\includegraphics[width=0.45\linewidth]{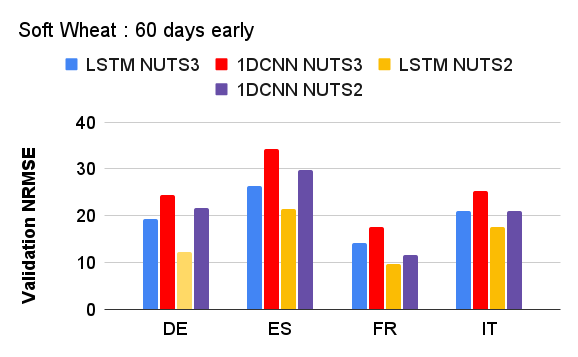}
\includegraphics[width=0.45\linewidth]{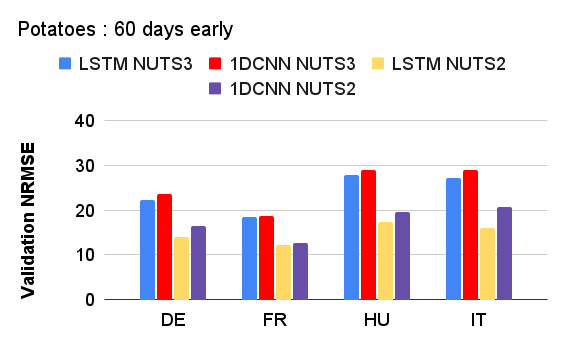}
\end{center}
\caption{Validation NRMSEs of weakly supervised NUTS3 and NUTS2 forecasts.}
\end{figure}

\subsection{Supplementary Results}
Wilcoxon signed rank test was run using prediction residuals (predicted yield - reported yield).

\begin{table}[!ht]
\caption{Wilcoxon-signed rank test (NUTS3)}
\label{wilcoxon-nuts3}
\begin{center}
\begin{tabular} { | p{0.25\linewidth} | p{0.2\linewidth} | p{0.2\linewidth} |}
\multicolumn{1}{c}{\bf MODEL}  &\multicolumn{1}{c}{\bf P-VALUE} &\multicolumn{1}{c}{\bf MEDIAN}
\\ \hline
 & \textbf{Soft wheat} & \\ \hline
WS LSTM & -- & -0.144 \\ \hline
Naive Trend & 0 & 0 \\ \hline
NUTS3 GBDT & 0.67 & -0.193 \\ \hline
NUTS3 Trend & 0 & 0.01 \\ \hline
 &  & \\ \hline
 & \textbf{Potatoes} & \\ \hline
WS LSTM & -- & 0.447 \\ \hline
Naive Trend & 0.00002 & -0.092 \\ \hline
NUTS3 GBDT & 0.265 & 0.397 \\ \hline
NUTS3 Trend & 0.07856 & 0.039 \\ \hline
\end{tabular}
\end{center}
\end{table}

\begin{table}[!ht]
\caption{Wilcoxon-signed rank test (NUTS2)}
\label{wilcoxon-nuts2}
\begin{center}
\begin{tabular} { | p{0.25\linewidth} | p{0.2\linewidth} | p{0.2\linewidth} |}
\multicolumn{1}{c}{\bf MODEL}  &\multicolumn{1}{c}{\bf P-VALUE} &\multicolumn{1}{c}{\bf MEDIAN}
\\ \hline
 & \textbf{Soft wheat} & \\ \hline
WS LSTM & -- & -0.089 \\ \hline
NUTS2 GBDT & 0.08346 & -0.0942 \\ \hline
NUTS2 Trend & 0.00095 & -0.005 \\ \hline
 &  & \\ \hline
 & \textbf{Potatoes} & \\ \hline
WS LSTM & -- & 0.494 \\ \hline
NUTS2 GBDT & 0.455 & 0.664 \\ \hline
NUTS2 Trend & 0.000002 & -0.1 \\ \hline
\end{tabular}
\end{center}
\end{table}

\begin{figure}[h]
\begin{center}
\includegraphics[width=0.45\linewidth]{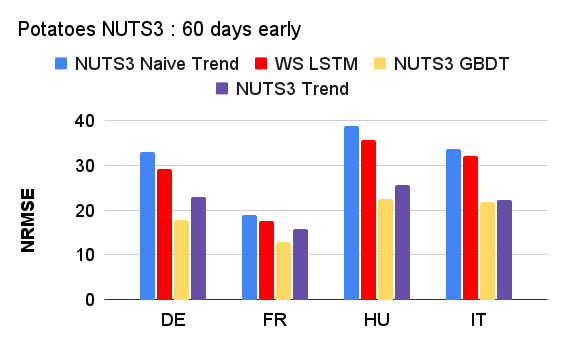}
\includegraphics[width=0.45\linewidth]{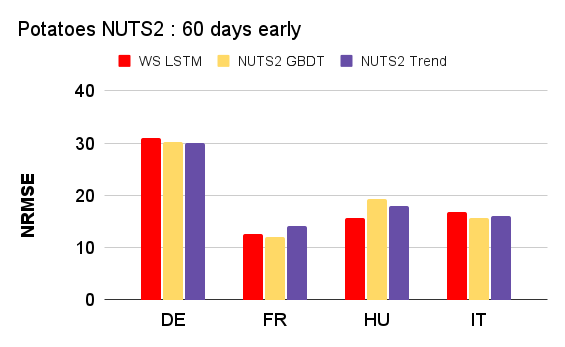}
\end{center}
\caption{Per-country NRMSEs for potatoes. \textit{Left}: NUTS3. \textit{Right}: NUTS2.} 
\end{figure}

\end{document}

%% file: math_commands.tex
\usepackage{amsmath,amsfonts,bm}

\def\eqref#1{equation~\ref{#1}}

\def\1{\bm{1}}

\DeclareMathAlphabet{\mathsfit}{\encodingdefault}{\sfdefault}{m}{sl}
\SetMathAlphabet{\mathsfit}{bold}{\encodingdefault}{\sfdefault}{bx}{n}

